\documentclass[letterpaper]{article} 
\usepackage{aaai23}  
\usepackage{times}  
\usepackage{helvet}  
\usepackage{courier}  
\usepackage[hyphens]{url}  
\usepackage{graphicx} 
\usepackage{natbib}  
\usepackage{caption} 
\usepackage{algorithm}
\usepackage{algorithmic}

\usepackage{xcolor}

\usepackage{newfloat}
\usepackage{listings}
\DeclareCaptionStyle{ruled}{labelfont=normalfont,labelsep=colon,strut=off} 
\floatstyle{ruled}
\newfloat{listing}{tb}{lst}{}
\floatname{listing}{Listing}
\title{FineDeb: A Debiasing Framework for Language Models}
\author{Akash Saravanan\equalcontrib, Dhruv Mullick\equalcontrib, Habibur Rahman\equalcontrib, Nidhi Hegde}
\affiliations{\textit{\{asaravan, mullick, habibur, nidhi.hegde\}@ualberta.ca}\\ Department of Computing Science, Alberta Machine Intelligence Institute (Amii) \\
University of Alberta, Canada}

\usepackage{bibentry}

\newcommand{\dhruv}[1]{\textcolor{black}{#1}}
\newcommand{\akash}[1]{\textcolor{black}{#1}}
\newcommand{\habib}[1]{\textcolor{black}{#1}}

\begin{document}

\maketitle

\begin{abstract}
As language models are increasingly included in human-facing machine learning tools, bias against demographic subgroups has gained attention.  
We propose FineDeb, a two-phase debiasing framework for language models that starts with contextual debiasing of embeddings learned by pretrained language models. The model is then fine-tuned on a language modeling objective. 
Our results show that FineDeb offers stronger debiasing in comparison to other methods which often result in models as biased as the original language model.
Our framework is generalizable for demographics with multiple classes, and we demonstrate its effectiveness through extensive experiments and comparisons with state of the art techniques. We release our code and data on GitHub \footnote{\url{https://github.com/akashsara/debiasing-language-models}}.
\end{abstract}

\section{Introduction \& Related Work}



Machine learning tools that rely on natural language processing (NLP) are increasingly being developed for scenarios with immediate impact on individuals, such as healthcare \citep{velupillai2018-healthcare}, conversational agents \citep{zhang-etal-2020-dialogpt}, and legal systems ~\citep{dale2019-law}. While the language models here vary in the type of embeddings used, they rely on representations that may reflect or exhibit bias \cite{manzini2019black, bolukbasi2016embeddings}. When used in downstream tasks such as prediction, health care diagnoses, or other decision-making, such representations can amplify bias and result in discriminatory actions against individuals in disadvantaged demographic subgroups. Our focus in the present paper is on such representational harms~\cite{blodgett2020language}. 

Prior work studies word embeddings to analyze bias and proposes debiasing techniques for NLP methods. 
Debiasing on word embeddings was first introduced by \citet{bolukbasi2016embeddings} and further refined to enable debiasing on multiple classes by \citet{manzini2019black}. However, recent advances in NLP have been focused on large pretrained transformer-based language models (LM) such as BERT and GPT.  Such models differ in that rather than considering individual word embeddings, they create representations that take into account large and connected components such as sentences and context. 
For 
comprehensive bias mitigation we must consider bias in the context of sentences, beyond mere word embeddings.  We therefore focus on bias in transformer-based contextual LMs.

Previous work \citep{bolukbasi2016embeddings, liang2020sentence, liang2021-towards} applies debiasing techniques on embeddings after bias subspaces in these representations have been detected.  While such techniques attempt to create debiased representations, downstream tasks may not necessarily reflect debiased language.  Furthermore, good performance on those downstream tasks, \akash{such as entity extraction or text classification}, must still be an objective. \citet{bordia2019identifying} 
thus act on the training objective by adding a regularizer for debiasing. \akash{In place of a regularizer, we} use an entirely new training objective to minimize distance between relevant embeddings. 



A further limitation of state of the art debiasing techniques is that they largely consider demographics with binary classes or when they consider multiple classes, they focus on the three largest subgroups \cite{manzini2019black}.  In real scenarios, social disadvantage is represented through more than the binary or majority/minority dichotomy, with demographic groups containing many classes.  
We apply our debiasing methodology on demographics with multiple classes.
\dhruv{We propose a debiasing framework - FineDeb for large pretrained LMs, with two phases: one for debiasing the embeddings, and one for fine-tuning the language modeling head. \textbf{1)}} In the debiasing phase we modify representations by using a training objective that minimizes the distance between embeddings of target words while considering their sentence contexts. \textbf{2)} In \akash{the} second phase, we fine-tune the model \dhruv{on a language modeling task.} 

\akash{Our results show that the FineDeb framework applies a significantly stronger level of debiasing in comparison to other methods while slightly compromising on downstream language modeling performance.}
\dhruv{\akash{In this paper we extend} prior work in several ways: compiling lists of target words for three demographics, generating a debiasing dataset for future benchmarking, providing a framework for debiasing language models rather than directly modifying representations, demonstrating a debiasing \akash{training} objective, and allowing \akash{for the} debiasing of multiple classes.}

\section{Data}
We use two types of data - one for the debiasing phase and one for the LM finetuning phase.  

\subsection{Debiasing Data} \label{debiasing-data}
Our methodology starts by first finetuning the LM with a new objective function on a debiasing dataset, which consists of examples of debiased sentences. The debiased sentence examples are created using  word lists crafted with multiple classes per demographic (in english language).  
Our final debiased model is thus trained by debiasing for multiple classes.  Note that this is a novel contribution beyond word lists with pairs created previously.   




Our word lists are compiled from various sources, both online (\ref{appendix:b}) and existing work \citep{bolukbasi2016embeddings, zhao-learning-embeddings}, to create a list of target word tuples for each demographic (race, religion and gender). The word list contains 2 classes for gender\footnote{NLP datasets have limited data for other gender classes}, 5 classes for race, and 7 classes for religion. For each demographic, the word list consists of several tuples of target words, where within each tuple the words are comparable. For example, \textit{("Muslim", "Christian", "Jewish", "Hindu", "Buddhist", "Confucianist", "Taoist")} is one tuple within the religion word list. We compile 10 word tuples for Race, 32 for Religion and 158 for Gender. We make the word lists freely available in our codebase, providing samples in Appendix \ref{appendix:b}.

\label{sentence-pairs}
In order to generate our final debiasing dataset, we first craft sentence templates from the RedditBias \citep{barikeri-etal-2021-redditbias} dataset. RedditBias is a dataset of human conversation data from Reddit across four demographics: \textit{gender, race, religion,} and \textit{queerness}. We convert all sentences containing the former three demographics into sentence templates by removing the target words. An example for race would be \textit{"all \_\_\_\_ are criminals"}. During the training process we choose a relevant word tuple and sample target words from different classes such as \textit{Black} and \textit{White} to generate sentence pairs that differ in only the target word.  We generate such sentence sets pairwise among all classes. 

\subsection{LM finetuning Data} \label{finetune_data}
Taking inspiration from prior work \citep{qian2019reducing, bordia2019identifying}, we use CNN-DailyMail \citep{hermann-2015-teaching} for our language model finetuning objective. It consists of $300,000+$ English news articles. 

\section{FineDeb}
Our method, FineDeb, adopts a two phase approach for training. In the first phase, we debias the model by modifying the embeddings learned by the language model, and in the second phase, we finetune the debiased model on the language modeling objective. \dhruv{We evaluate our framework in terms of bias and downstream language modeling performance using standard metrics.}
Our method is demonstrated on a BERT model \cite{devlin-2019-bert}, specifically \textit{bert-base-uncased}. All hyperparameters are listed in Appendix \ref{sec:appendix_hyperpams}.

\subsection{Debiasing Phase} \label{debias_phase}
In the debiasing phase, we train our model using the sentence pairs generated in Section \ref{sentence-pairs}. Most modern language models are contextual in nature \cite{devlin-2019-bert}, that is, the same word may have different meanings based on the context in which the word is used. For instance, the word \textit{"temple"} could refer to a building or a part of the human body. Thus it becomes important to perform our debiasing only in those contexts where the bias may exist. 
Our methodology is inspired by the traditional process of determining relationships between pairs of words by computing the distance between their embeddings \cite{mikolov-2013-word2vec,bolukbasi2016embeddings}. However, unlike these methods, we account for the contextual meaning of the sentence.

Given two near-identical sentences that differ only by a target word (or phrase), we first compute the difference between the embeddings of the two sentences (using their [CLS] token embedding as in \citet{devlin-2019-bert}), and the difference between the embeddings of the target words in each sentence. Our training objective is motivated by the idea that since the two sentences differ only in their target words, logically, the difference in the embeddings of these two sentences should also differ only by the difference in the embeddings of the target words. Our training objective thus minimizes the distance between these two quantities to debias the model. Formally, our loss function is as follows: $L(S_1, S_2, W_1, W_2) = \mathcal{D}(S_1 - S_2, W_1 - W_2),$
where $S_i$ is the embedding for sentence $i$, $W_i$ is the embedding for the target word in sentence $i$, 
and $\mathcal{D}(\cdot)$ is the distance between the two quantities (we use the Mean Squared Error but other similarity measures such as Cosine similarity can also be used). Taking an example from \citet{nangia-etal-2020-crows}, if we had two sentences \textit{"The crafty Jews made a plan to steal the money."} and \textit{"The crafty Christians made a plan to steal the money."}, $W_1$ and $W_2$ would be the embeddings of \textit{"Jews"} and \textit{"Christians"} respectively, while $S_1$ and $S_2$ would respectively be the sentence embeddings.

\subsection{LM Finetuning Phase} \label{lm_fine_tune_phase}

Pretrained transformer models consist of an embedding-generating model, and a language modeling head (LM head) which gives probabilities for each of the words based on these embeddings. In the case of our debiasing phase, we only update the embedding-generating model and not the LM head. We then finetune the entire model (debiased model + LM head) on the standard BERT language modeling objective (masked word prediction) using the CNN-DailyMail dataset.

During this finetuning process, we freeze the debiased model so that the debiased embeddings do not change, and only the weights in the LM head are updated. \dhruv{This results in an overall model suitable for the LM task}. While this method of improving the language modeling ability of our model may re-introduce some biases that exist in the CNN-DailyMail dataset, it does not eliminate the effects of our debiasing, as we show in our results. 
Indeed the two phases may counteract each other, but this tradeoff can be controlled via the amount of training done in each phase.

\section{Metrics}\label{sec:metrics}

We evaluate FineDeb on the three demographics of gender, race, and religion using three metrics: StereoSet, SEAT, and Crow-S Pairs. We choose these metrics due to their popularity in prior work \cite{liang2021-towards,survey_debiasing_ptlm_2021}, though we note that other metrics may exist \cite{kurita-etal-2019-measuring}. These metrics differ in how bias is evaluated, the data used in evaluating bias, and whether the language modeling performance is considered.

\subsection{StereoSet}
Following recent work \cite{survey_debiasing_ptlm_2021, opt_2022}, we use StereoSet \cite{nadeem-etal-2021-stereoset} to evaluate our work. StereoSet measures the Stereotype Score (SS) which gives a measure of bias 
and the Language Modeling Score (LMS) which determines performace 
at language modeling tasks. There is a trade-off here, 
as a model could be perfectly unbiased but be a poor language model (or vice versa). Thus the authors provide a combined measure named ICAT. Following prior work \citep{survey_debiasing_ptlm_2021}, we use the intrasentence variant of StereoSet. 


\subsection{Crow-S Pairs  }
The Crowdsourced Stereotype Pairs (Crow-S Pairs) \cite{nangia-etal-2020-crows} uses crowdsourced pairs of sentences that differ only by a small number of tokens such that one sentence reflects a stereotype while the other violates that stereotype. 
Under this metric, a perfect model is equally likely to pick the stereotypical sentence as it is to pick the anti-stereotypical sentence. This metric does not test the language modeling ability of the model but covers a wide variety of biases. 

\subsection{SEAT}
Sentence Encoder Association Test (SEAT) \cite{may-etal-2019-measuring-seat} is a sentence level extension of the WEAT metric \cite{caliskan-weat-2017} which measures bias between two sets of attribute words and two sets of target words.  
Specifically, SEAT uses sentence templates to obtain representations of words. The metric is measured in terms of the average effect size across several tests, where a value closer to 0 indicates a lower degree of bias but it 
does not test the language modeling ability of the model.  

\section{Results and Discussion} \label{results}

\begin{table*}[ht]
\centering
\resizebox{1.95\columnwidth}{!}{
\begin{tabular}{lccc|ccc|ccc}
\hline
\multicolumn{1}{l}{} & \multicolumn{3}{c|}{\textbf{Gender}} & \multicolumn{3}{c|}{\textbf{Race}} & \multicolumn{3}{c}{\textbf{Religion}} \\
StereoSet & LMS & SS & ICAT & LMS &  SS &  ICAT  &  LMS &  SS & ICAT \\ \hline
BERT & 84.17 & 60.28 & 66.86 & \underline{84.17}	& 57.03	& 72.34	& 84.17	& 59.70 &	67.84  \\ 
BERT (LM finetuning)	& \textbf{85.01}	& 59.01	& \underline{69.69} &	83.78 &	56.38 &	73.08 &	83.07 &	61.47 &	64.01  \\ 
FineDeb &	77.70&\textbf{	53.27}&	\textbf{72.62}&	75.37&	\textbf{50.82}&	\underline{74.13} &	74.10&	\textbf{50.39}&	\textbf{73.52}  \\ 
CDA &	83.08 &	59.61 &	67.11 &	83.41 &	56.73 &	72.18	& 83.24 &	58.37 &	69.31  \\ 
Dropout &	83.04 &	60.66 &	65.34 &	83.04 &	57.07 &	71.30 &	83.04 &	59.13 &	67.88  \\ 
INLP &	80.63 &	\underline{57.25} &	68.94 &	83.12 &	57.29 &	71.00 &	83.36 &	60.31 &	66.17  \\ 
Self-Debias &	84.09 &	59.34 &	68.38 &	\textbf{84.24} &	\underline{54.30} &	\textbf{77.00} &	\underline{84.23} &	\underline{57.26} &	\underline{72.00}  \\ 
Sentence Debias	& \underline{84.20} & 	59.37 &	68.42 &	83.95 &	57.78 &	70.89 &	\textbf{84.26} &	58.73 &	69.55  \\ \hline
\end{tabular}
}
\centering
\caption{StereoSet evaluation. LMS indicates Language Modeling Score, SS is the Stereotype Score, ICAT is the overall score. Higher is better for LMS \& ICAT, closer to 50 is better for SS (\textbf{Best}; \underline{Next Best}).}
\label{table:stereoset}
\end{table*}

\begin{table}[ht]
\small

\centering
\resizebox{0.95\columnwidth}{!}{
\begin{tabular}{lccc}
\hline
\multicolumn{1}{l}{} & \multicolumn{1}{c}{\textbf{Gender}} & \multicolumn{1}{c}{\textbf{Race}} & \multicolumn{1}{c}{\textbf{Religion}} \\ \hline
BERT &	0.62&	0.62&	0.49 \\
FineDeb &	\underline{0.36}&	0.62&	0.67 \\
CDA&	0.72&	\underline{0.57}&	\textbf{0.34} \\
Dropout&	0.77&	\textbf{0.55}&	\underline{0.38}  \\
INLP&	\textbf{0.20}&	0.64&	0.46  \\
Sentence Debias&	0.43&	0.61&	0.44  \\ \hline
\end{tabular}
}
\centering
\caption{SEAT: Average effect size for each demographic. BERT (LM finetuning) \& Self-Debias do not modify internal representations and so have the same SEAT score as BERT. Lower is better (\textbf{Best}; \underline{Next Best}). }
\label{table:seat}
\end{table}

\begin{table}[!ht]
\centering
\centering
\resizebox{0.95\columnwidth}{!}{
\begin{tabular}{lccc}
\hline
\multicolumn{1}{l}{} & \multicolumn{1}{c}{\textbf{Gender}} & \multicolumn{1}{c}{\textbf{Race}} & \multicolumn{1}{c}{\textbf{Religion}} \\ \hline
BERT &	57.25 &	62.33 &	62.86 \\
BERT (LM finetuning)	& 57.63	& 62.91 &	58.10 \\
FineDeb&	54.58  &	65.24 &	\textbf{44.76} \\
CDA&	56.11 &	\textbf{56.70 }&	60.00   \\
Dropout&	55.34&	\underline{59.03}&	\textbf{55.24}  \\
INLP&	\textbf{51.15}&	67.96 &	60.95   \\
Self-Debias&	\underline{52.29}&	\textbf{56.70}&	\underline{56.19}  \\ 

Sentence Debias &	\underline{52.29}&	62.72&	63.81 \\ \hline
\end{tabular}
}
\caption{Crow-S Pairs: Metric scores for each demographic. Closer to 50 is better (\textbf{Best}; \underline{Next Best}).}

\label{table:crow}
\end{table}

We compare FineDeb on three demographics 
against two baselines and five prior works. Our baselines are pretrained BERT and a pretrained BERT model where only the LM finetuning phase is applied. We include results on this latter model to show that the LM finetuning phase does not significantly alter bias in the model compared to the base BERT model, and any change in bias in our model is strictly due to our debiasing phase. This is evident in the results which show that BERT with LM finetuning is on par with or slightly better than the base BERT model for all listed metrics. 
We also compare with state of the art methods - CDA, Dropout, INLP, Self-Debias and, Sentence Debias, the results of which we cite from \citet{survey_debiasing_ptlm_2021}.

We first present results\footnote{Evaluation code taken from \citet{nadeem-etal-2021-stereoset} (StereoSet) and \citet{survey_debiasing_ptlm_2021} (SEAT and Crow-S)} on the StereoSet metric in Table~\ref{table:stereoset}. Under the SS measure, FineDeb outperforms all techniques for all three demographics. 
Furthermore, the other debiasing methods have small improvements in comparison to the baseline BERT model while our work shows near-perfect scores for two of the three demographics (race and religion).
Under the LMS measure, Self-Debias, Sentence Debias, and the baselines have the best or second best results, across different demographics.  It is expected that our method not perform best on the LMS measure since we focus first and foremost on debiasing.  However the reduction in LMS performance is not too severe since under the combined measure of ICAT our model performs best for gender and religion, and second best for race.

For SEAT (Table~\ref{table:seat}), our method performs second best for gender, while for race and religion we do better than one method and no other methods respectively.  We note that SEAT measures distances between embeddings, whereas StereoSet is based on final word outputs. Further, SEAT does not measure LM performance.  We reason that since disadvantage and 
discriminatory harm is manifested due to contextual outputs from language models and not just their representations, a metric such as StereoSet, which considers both a measure of final LM performance, as well as a measure of the bias in contextual outputs, more accurately represents real-world impact of bias.

The results for Crow-S (Table~\ref{table:crow}) show that, FineDeb outperforms all others, with similar performance as Dropout \habib{for religion}. For gender, our method beats the baseline and two other methods, but performs poorly for race. Crow-S measures bias similarly to StereoSet by considering whether a sterotypical sentence is preferred among sentence pairs. The sentences in a pair, however, differ on attributes rather than target words. We cannot be certain if this difference accounts for the differing results, but we note that \habib{like SEAT,} Crow-S does not measure LM ability, an important component of any model.

\section{Conclusions and Future Work}

In this paper, we have proposed FineDeb, a new framework for debiasing language models and have demonstrated its performance on commonly used evaluation metrics. Considering the results over three demographics and across three metrics against all benchmarks, we see that our FineDeb framework is strongly debiasing.
This could be since the training objective itself minimizes embedding distance. \akash{At the same time, this may lead to a distortion in the embeddings}. This is seen in the near-perfect StereoSet SS measures and in the somewhat poorer scores for SEAT which directly works with the embeddings. 
Since the other debiasing solutions have an SS score similar to the default BERT case, their debiasing ability is not as strong as FineDeb. 
For the ICAT measure which combines debiasing and LM performance our method performs the best or second best, suggesting a better overall performance for our method. 

The tradeoff between the debiasing performance (measured by SS) and language modeling performance (measured by LMS score) which we note in Section \ref{lm_fine_tune_phase} can be controlled by hyperparameter tuning during training. A deteriorated language modeling ability would have an impact on other downstream NLP tasks \cite{wang-etal-2019-tell}. Depending on the use case, either a more debiased model or a model with better language modeling ability (and consequently better downstream NLP task ability) can be obtained.

We note that despite FineDeb achieving near-perfect debiasing performance at the cost of language modeling performance, the language modeling performance may not always be of primary importance. We envision a number of scenarios in which the extent of debiasing would be more important than LM performance. It is easier for a human to identify and check for correctness than it is to identify and correct biases, some of which they may not even be aware of. Thus, any human-in-the-loop task would benefit from a model generating suggestions free from biases. For instance, consider the case of generating text descriptions for products \citet{koto-etal-2022-pretrained}. A debiased model can first generate an unbiased description, and then a human can fix any errors, if necessary. Similarly FineDeb can play an important role in other human-in-the-loop scenarios like generating stories, or in recommendation systems where a slight decrease in recommendation quality is preferable to serving biased recommendations.


There are a few avenues for future work in this area. Namely, expanding our work to include more classes (such as in gender) or to other demographics; a more comprehensive analysis of our model on downstream tasks \akash{and modifying our current two-phase framework approach to instead use a single phase of interleaved debiasing and finetuning. Further, current metrics for bias rely on comparisons of either target words or attribute words, resulting in varying performance across the different techniques. This suggests the need for a more comprehensive metric on bias that is agnostic to the debiasing technique. }

\section{Limitations}

To encourage transparency in research, we disclose some limitations of our work in the hope that it offers further insights our the proposed framework:

\begin{itemize}
    \item There may be cases where we want a bias to exist. For example, in the sentence "The \_\_\_ man went to the mosque.", the probability of "Muslim" should be higher than the probability of "Christian" or "Jew". While people of any religion could go to a mosque, a person who follows Islam is far more likely. This falls under explainable bias \cite{mehrabi_2021}.
    \item While our method is effective, it relies heavily on the word lists that we have compiled to the best of our knowledge. These are by no means fully representative of all bias targets for a given demographic and there is still scope for expansion.
\end{itemize}

\appendix
\section*{Appendix}

\begin{table}[!ht]
    \centering
    \small
    \centering
    \centering
    \resizebox{0.6\columnwidth}{!}{
    \begin{tabular}{ll}
    \hline
        Male & Female \\ \hline
        countryman & countrywoman \\
        fraternal & sororal \\
        manservant & maidservant \\
        divo & diva \\
        actor & actress \\
        bachelor & spinster \\
        papa & mama \\
        busboy & busgirl \\ \hline
    \end{tabular}
    }
    \caption{Example word lists for Gender}
    \label{table:word_list_gender}
\end{table}

\begin{table*}[!ht]
    \centering
    \centering
    \resizebox{1.6\columnwidth}{!}{
    \begin{tabular}{lllll}
    \hline
        African-American & Anglo-American & Hispanic & Asian & Native-American \\ \hline
        Black & White & Latino & Brown & light-brown \\
        Negroid & Caucasian & Latino & Brown & Native-American \\
        African-American & Anglo-American & Hispanic & Asian & Native-American \\
        Afro-American & Anglo-American & Hispanic & Asian & Native-American \\
        African & American & Hispanic & Asian & Native-American \\
        Afroamerican & Angloamerican & Hispanic & Asian & Native-American \\
        Negro & Caucasian & Hispanic & Brown & Native-American \\
        dark-skin & light-skin & white-latino & gray-skin & reddish-brown \\
        dark-skin & light-skin & black-latino & gray-skin & reddish-brown \\ \hline
    \end{tabular}
    }
    \small
    \caption{Example word lists for Race}
    \label{table:word_list_race}
\end{table*}

\begin{table*}[!ht]
    \small
    \centering
    \centering
    \centering
    \resizebox{2\columnwidth}{!}{
    \begin{tabular}{lllllll}
    \hline
        Islam & Christianity & Judaism & Hinduism & Buddhism & Confucianism & Taoism \\ \hline
        Islam & Christianity & Judaism & Hinduism & Buddhism & Ruism & Daoism \\ 
        Quran & Bible & Torah & Gita & Tripitaka & Analects & Tao-Te-Ching \\ 
        Koran & Bible & Tanakh & Veda & Tripitaka & Analects & Tao-Te-Ching \\
        Muslim & Christian & Jewish & Hindu & Buddhist & Confucianist & Taoist \\ 
        islamic & Christian & Jewish & Hinduism & Buddhist & Confucius & Dao \\
        Mohammed & Jesus & Malachi & Ramakrishna & Gautama & Confucius & Laozi \\
        Mohammed & Jesus & Moses & Krishna & Gautama & Kung-Fu-Tzu & Laozi \\ \hline
    \end{tabular}
    }
    \caption{Example word lists for Religion}
    \label{table:word_list_religion}
\end{table*}

\begin{table*}[!ht]
    \centering
    \small
    \centering
    \centering
    \resizebox{2.0\columnwidth}{!}{
    \begin{tabular}{lcccccc|c}
    \hline
        \multicolumn{1}{c}{Model} &  \multicolumn{1}{c}{\textbf{SEAT-6}} & \multicolumn{1}{c}{\textbf{SEAT-6b}} & \multicolumn{1}{c}{\textbf{SEAT-7}} & \multicolumn{1}{c}{\textbf{SEAT-7b}} & \multicolumn{1}{c}{\textbf{SEAT-8}} & \multicolumn{1}{c}{\textbf{SEAT-8b}} & \multicolumn{1}{c}{\textbf{Avg. Effect Size}} \\\hline
        bert-base-uncased & \underline{0.931} & 0.090  & -0.124  & \underline{0.937}  & \underline{0.783}  & \underline{0.858} & 0.620 \\
        FineDeb & 0.248 & 0.113 & 0.179 & 0.298 & 0.280 & \underline{1.056} & 0.362 \\
        CDA  & \underline{.846}  & 0.186  & -0.278  & \underline{1.342}  & \underline{0.831}  & \underline{0.849} & 0.722 \\
        Dropout  & \underline{1.136} & 0.317 & 0.138 & \underline{1.179} & \underline{0.879} & \underline{0.939} & 0.765 \\
        INLP  & 0.317  &  -0.354  & -0.258  & 0.105 & 0.187  & -0.004 & 0.204 \\
        Sentence Debias & 0.350 &-0.298 &-0.626 & \underline{0.458} & 0.413 & \underline{0.462} & 0.434 \\ \hline
    \end{tabular}
    }
    \centering
    \caption{SEAT effect sizes (closer to 0 is better) for Gender debiased models. \underline{Underlined} if statistically significant $(p < 0.01)$.}
    \label{table:detailed_seat_gender}
\end{table*}

\begin{table*}[!ht]
    \centering
    \small
    \centering
    \centering
    \resizebox{2.0\columnwidth}{!}{
    \begin{tabular}{lccccccc|c}
    \hline
        \multicolumn{1}{c}{Model} &  \multicolumn{1}{c}{\textbf{ABW-1}} & \multicolumn{1}{c}{\textbf{ABW-2}} & \multicolumn{1}{c}{\textbf{SEAT-7}} & \multicolumn{1}{c}{\textbf{SEAT-3}} & \multicolumn{1}{c}{\textbf{SEAT-4}} & \multicolumn{1}{c}{\textbf{SEAT-5}} & \multicolumn{1}{c}{\textbf{SEAT-5b}} & \multicolumn{1}{c}{\textbf{Avg. Effect Size}} \\ \hline
        bert-base-uncased & -0.079 & \underline{0.690} & \underline{0.778} & \underline{0.469} & \underline{0.901} & \underline{0.887} & \underline{0.539} & 0.620 \\
        FineDeb & \underline{1.111}	& -0.235 &	\underline{0.806} &	0.085 &	\underline{0.858} &	\underline{0.787} &	\underline{0.456} &	0.620 \\
        CDA  & 0.231  & \underline{0.619}  & \underline{0.824}  & \underline{0.510}  & \underline{0.896}  & \underline{0.418}  & \underline{0.486} & 0.569 \\
        Dropout  & \underline{0.415}  & \underline{0.690}  & \underline{0.698}  & \underline{0.476}  & \underline{0.683}  & \underline{0.417}  & \underline{0.495} & 0.554 \\
        INLP  & 0.295  & \underline{0.565}  & \underline{0.799}  & \underline{0.370}  & \underline{0.976}   & \underline{1.039}  & \underline{0.432} & 0.639 \\
        Sentence Debias  & -0.067  & \underline{0.684}  & \underline{0.776}  & \underline{0.451}  & \underline{0.902}  & \underline{0.891}  & \underline{0.513} & 0.612\\ \hline
    \end{tabular}
    }
    \caption{SEAT effect sizes (closer to 0 is better) for Race debiased models. \underline{Underlined} if statistically significant $(p < 0.01)$.}
    \label{table:detailed_seat_race}
\end{table*}

\begin{table*}[!ht]
    \centering
    \small
    \centering
    \centering
    \resizebox{1.6\columnwidth}{!}{
    \begin{tabular}{lcccc|c}
    \hline
        \multicolumn{1}{c}{Model} &  \multicolumn{1}{c}{\textbf{Religion-1}} & \multicolumn{1}{c}{\textbf{Religion-1b}} & \multicolumn{1}{c}{\textbf{Religion-2}} & \multicolumn{1}{c}{\textbf{Religion-2b}} &\multicolumn{1}{c}{\textbf{Avg. Effect Size}} \\\hline
        bert-base-uncased  & \underline{0.744}  & -0.067  & \underline{1.009}  & -0.147 & 0.492 \\
        FineDeb &	\underline{0.697} &	\underline{0.701}	&	\underline{0.666} &	\underline{0.613} &	0.670 \\
        CDA  & 0.355  & -0.104  & \underline{0.424}  & -0.474 & 0.339 \\
        Dropout  & \underline{0.535}  & 0.109  & \underline{0.436}  & -0.428 & 0.377 \\
        INLP  & \underline{0.473}  & -0.301  & \underline{0.787}  & -0.280 & 0.460 \\
        Sentence Debias  & \underline{0.728}  & 0.003  & \underline{0.985}  & 0.038 & 0.439 \\ \hline
    \end{tabular}
    }
    \caption{SEAT effect sizes (closer to 0 is better) for Religion debiased models. \underline{Underlined} if statistically significant $(p < 0.01)$.}
    \label{table:detailed_seat_religion}
\end{table*}

\section{Word Lists}
\label{appendix:b}
We compile our word lists from several online sources in addition to prior work \cite{bolukbasi2016embeddings, zhao-learning-embeddings}. Specifically, we consider the following pages on Wikipedia: Major Religious Groups, Race (Human Categorization). Im addition, we also consulted https://courses.lumenlearning.com/sociology/chapter/world-religions/.

\subsection{Gender}
In Table \ref{table:word_list_gender}, we provide a sample of word lists in the Gender word list.

\subsection{Race}
In Table \ref{table:word_list_race}, we provide a sample of word lists in the Race word list.

\subsection{Religion}
In Table \ref{table:word_list_religion}, we provide a sample of word lists in the Religion word list.

\section{Additional Results}

We provide the complete evaluation results for SEAT in Tables \ref{table:detailed_seat_gender}, \ref{table:detailed_seat_race} and \ref{table:detailed_seat_religion}. 

\section{Training Details}
For each model, we run a job with a time limit of 41 hours, 4 CPUs per job, 4GB of memory per CPU and 1 NVIDIA V100 GPU.

\section{Hyperparameters}
\label{sec:appendix_hyperpams}

Considering the trade-off between bias reduction and LM performance discussed in this paper, we experiment with different dataset sizes for both the debiasing phase and LM finetuning phase. Intuitively, a larger debiasing dataset size would lead to a higher bias reduction but poorer LM performance. On the other hand, a large LM finetuning dataset would lead to higher bias but also better LM performance. The debiasing dataset sampling size is varied as - 500, 1000, 2000, 4000, and continues upto the debiasing dataset's size (different for different demographics). The LM finetuning dataset size is varied as a percentage of the total CNN dataset - 1\%, 2\%, 4\%, 8\%, 16\% and 32\%.

For training our language model (\ref{debias_phase}, \ref{lm_fine_tune_phase}) we use a batch size of 64 to maximise GPU usage, maximum source length of 64. For language model finetuning, we use 100 epochs to allow for convergence, whereas for debiasing we find 30 epochs to be sufficient, with both training processes utilizing an early stopping mechanism. We use a learning rate of 1e-4 based on what we find in the original BERT paper \cite{devlin-2019-bert}.

\bibliography{aaai23,anthology,custom}

\end{document}